\PassOptionsToPackage{table}{xcolor}
\RequirePackage{fix-cm}
\documentclass{style}

\usepackage[utf8]{inputenc}

\usepackage{amsmath,amsfonts,bm}

\def\eqref#1{equation~\ref{#1}}

\def\1{\bm{1}}

\DeclareMathAlphabet{\mathsfit}{\encodingdefault}{\sfdefault}{m}{sl}
\SetMathAlphabet{\mathsfit}{bold}{\encodingdefault}{\sfdefault}{bx}{n}

\usepackage{amsmath,amssymb}
\usepackage{url}
\usepackage{graphicx}
\usepackage{capt-of}
\usepackage{wrapfig}
\usepackage{enumitem,float,nicefrac}
\usepackage{needspace}
\title{G-STAR: Group-Scaled Token Capacity for Deeply Compressed Autoregressive Generation}
\author{%
  \mbox{Xu Huang\texorpdfstring{$^{1,2*}$}{}}\hspace{0.14cm}
  \mbox{Ye Huang\texorpdfstring{$^{1*}$}{}}\hspace{0.14cm}
  \mbox{Xiaotong Li\texorpdfstring{$^{2\dagger\ddagger}$}{}}\hspace{0.14cm}
  \mbox{Zhengrong Yue\texorpdfstring{$^{2,3}$}{}}\hspace{0.14cm}
  \mbox{Jialv Zou\texorpdfstring{$^{2,4}$}{}}\hspace{0.14cm}
  \mbox{Minhao Jing\texorpdfstring{$^{2}$}{}}\\[0.35em]
  \mbox{Haodong Li\texorpdfstring{$^{2,5}$}{}}\hspace{0.14cm}
  \mbox{Wenhao Yan\texorpdfstring{$^{2,6}$}{}}\hspace{0.14cm}
  \mbox{Yaxin Luo\texorpdfstring{$^{2,7}$}{}}\hspace{0.14cm}
  \mbox{Duomin Wang\texorpdfstring{$^{1}$}{}}\hspace{0.14cm}
  \mbox{Daquan Zhou\texorpdfstring{$^{1\ddagger}$}{}}
}
\affiliation[1]{Peking University}
\affiliation[2]{Meituan Longcat}
\affiliation[3]{Shanghai Jiao Tong University}
\affiliation[4]{Huazhong University of Science and Technology}
\affiliation[5]{The Chinese University of Hong Kong}
\affiliation[6]{Tsinghua University}
\affiliation[7]{MBZUAI}
\contribution[*]{Equal contribution}
\contribution[\dagger]{Project lead}
\contribution[\ddagger]{Corresponding author}
\abstract{
Discrete autoregressive models offer a scalable framework for image generation. However, they face a fundamental trade-off: long token sequences preserve representation quality but slow generation, while aggressive spatial compression reduces representation capacity. Based on product quantization, we show that, under a fixed nominal information budget, group size should scale quadratically with the per-axis spatial compression factor. Building on this principle, we propose GSTAR, a spatial--group co-scaling framework that enables \(32\times\) and even effective \(64\times\) compression while preserving strong reconstruction, understanding, and generation performance. Since larger groups introduce intra-group dependencies that must be modeled efficiently, we analyze their impact on inference latency and generation quality and introduce a semi-autoregressive decoder that combines parallel feature computation with lightweight causal modeling. With these designs, our highly compressed models perform competitively with leading models across five representative text-to-image benchmarks (GenEval, DPG-Bench, PRISM, LongText-Bench, and Qwen-Image-Bench). At \(1024\times1024\) resolution, our model delivers a \(20\times\) speedup over prior discrete autoregressive models, generating each image in 2.5 seconds on a single NVIDIA H200. Together, these results show that highly compressed discrete representations can enable high-quality autoregressive generation at low latency.
}
\newcommand{\linkicon}[1]{\raisebox{-0.22em}{\includegraphics[height=1.15em]{resources/#1}}\hspace{0.35em}}
\makeatletter
\renewcommand\checkdata[2][]{\addtolist[#1]{#2}{\checkdatalist}{\checkdataformat}{\par\vskip 2.2mm}}
\makeatother
\checkdata[\linkicon{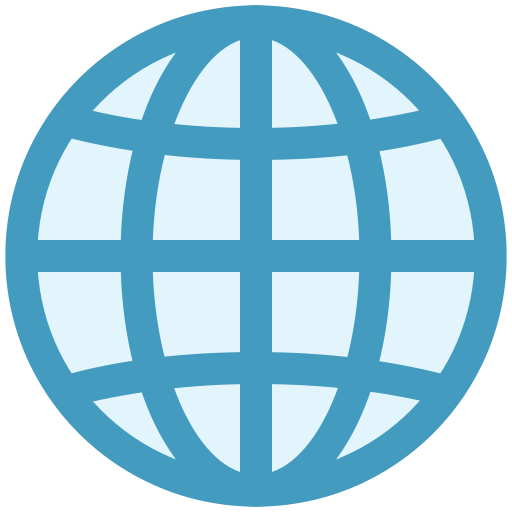}Project Page]{\href{https://dagroup-pku.github.io/G-STAR}{\texttt{https://dagroup-pku.github.io/G-STAR}}}
\checkdata[\linkicon{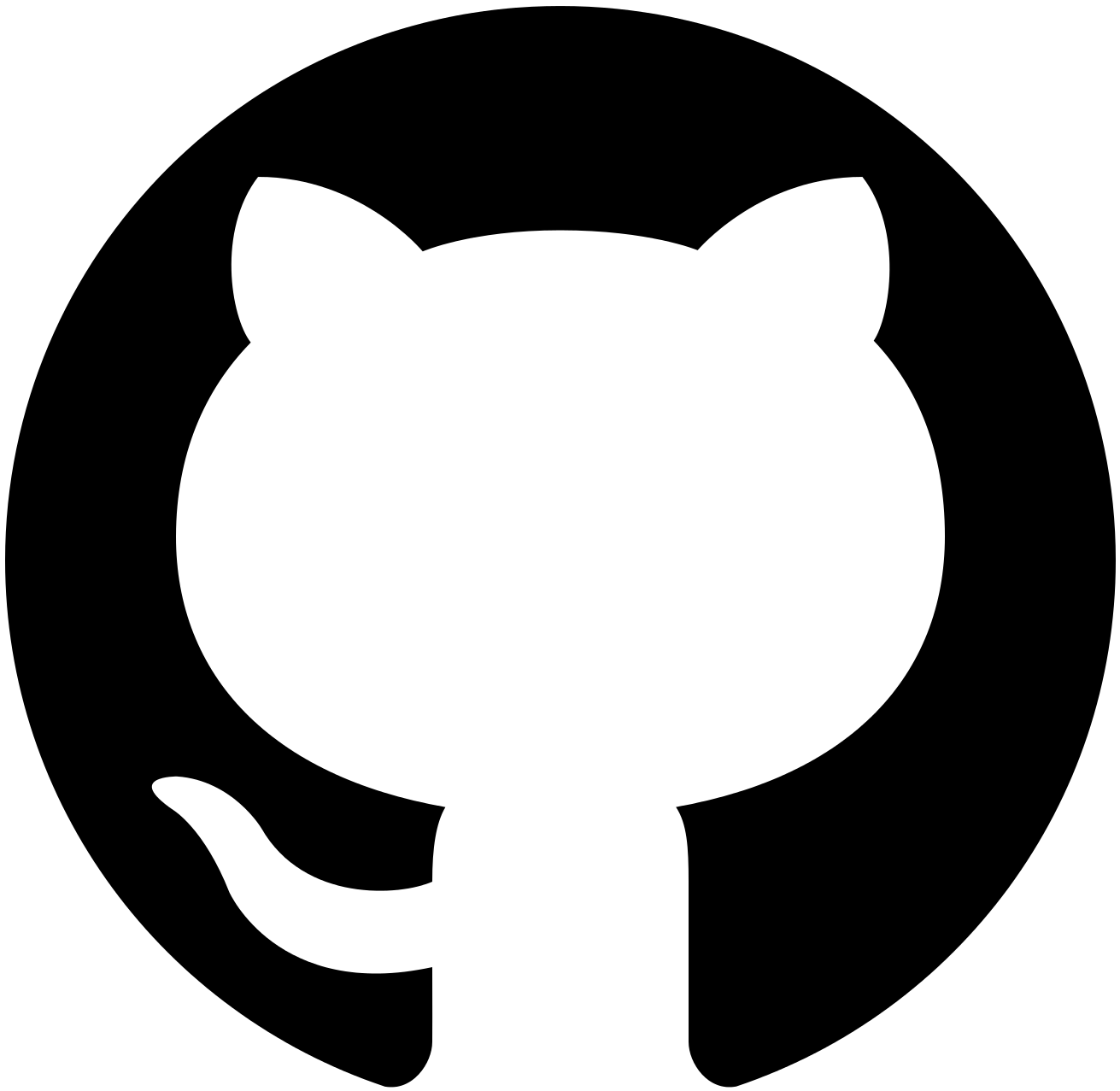}GitHub Repo]{\href{https://github.com/DAGroup-PKU/G-STAR}{\texttt{https://github.com/DAGroup-PKU/G-STAR}}}
\checkdata[\linkicon{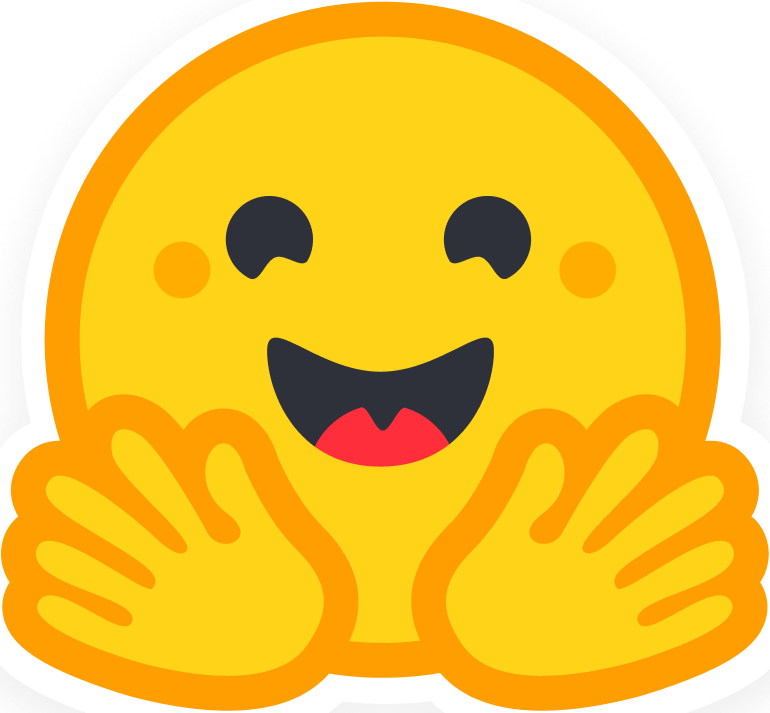}Huggingface Repo]{\href{https://huggingface.co/DAGroup-PKU/G-star}{\texttt{https://huggingface.co/DAGroup-PKU/G-star}}}
\hypersetup{
  pdftitle={GSTAR: Group-Scaled Token Capacity for Deeply Compressed Autoregressive Generation},
  pdfauthor={Xu Huang, Ye Huang, Xiaotong Li, Zhengrong Yue, Jialv Zou, Minhao Jing, Haodong Li, Wenhao Yan, Yaxin Luo, Duomin Wang, Daquan Zhou}
}

\makeatletter
\renewcommand{\authorfont}{\fontsize{11}{13.5}\selectfont}
\fancypagestyle{firststyle}{\fancyhf{}\fancyfoot[C]{\thepage}}
\renewcommand{\@toptitlebar}{%
  \vspace*{-14mm}%
  \noindent\includegraphics[height=10mm]{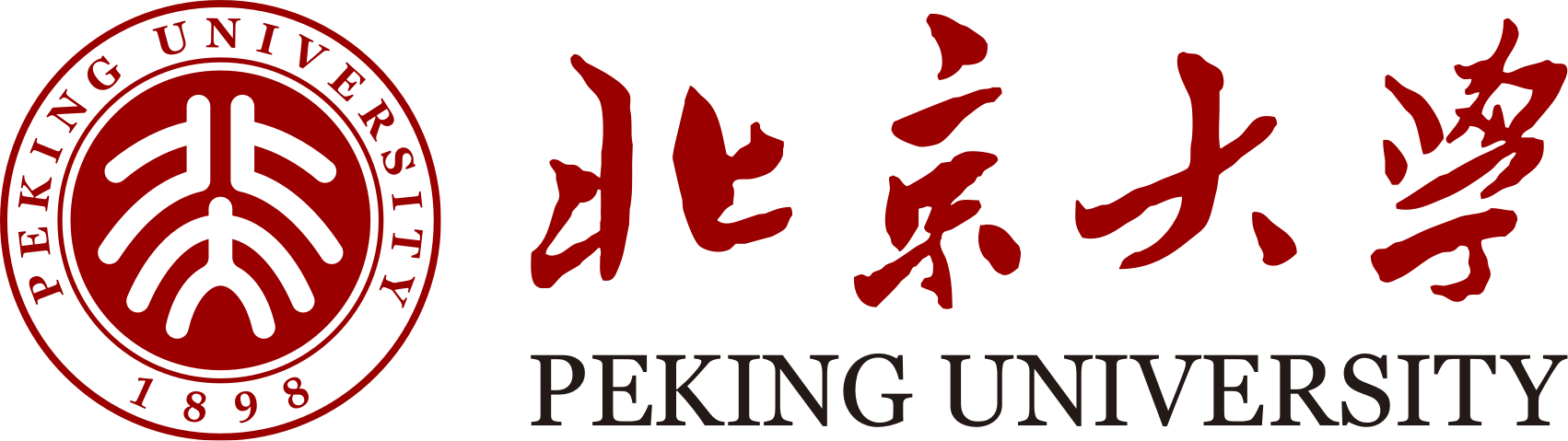}\hfill\includegraphics[height=7.5mm]{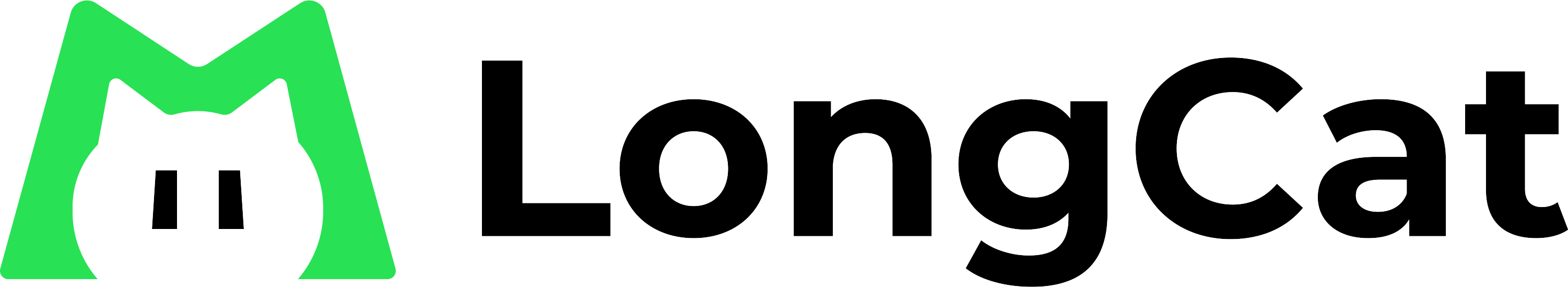}\par
  \vskip 1.2mm
  \pkured{\hrule height 2.5pt}
  \vskip 6mm
}
\makeatother

\begin{document}
\maketitle

\begin{figure*}[p]
  \centering
  \includegraphics[width=0.85\textwidth]{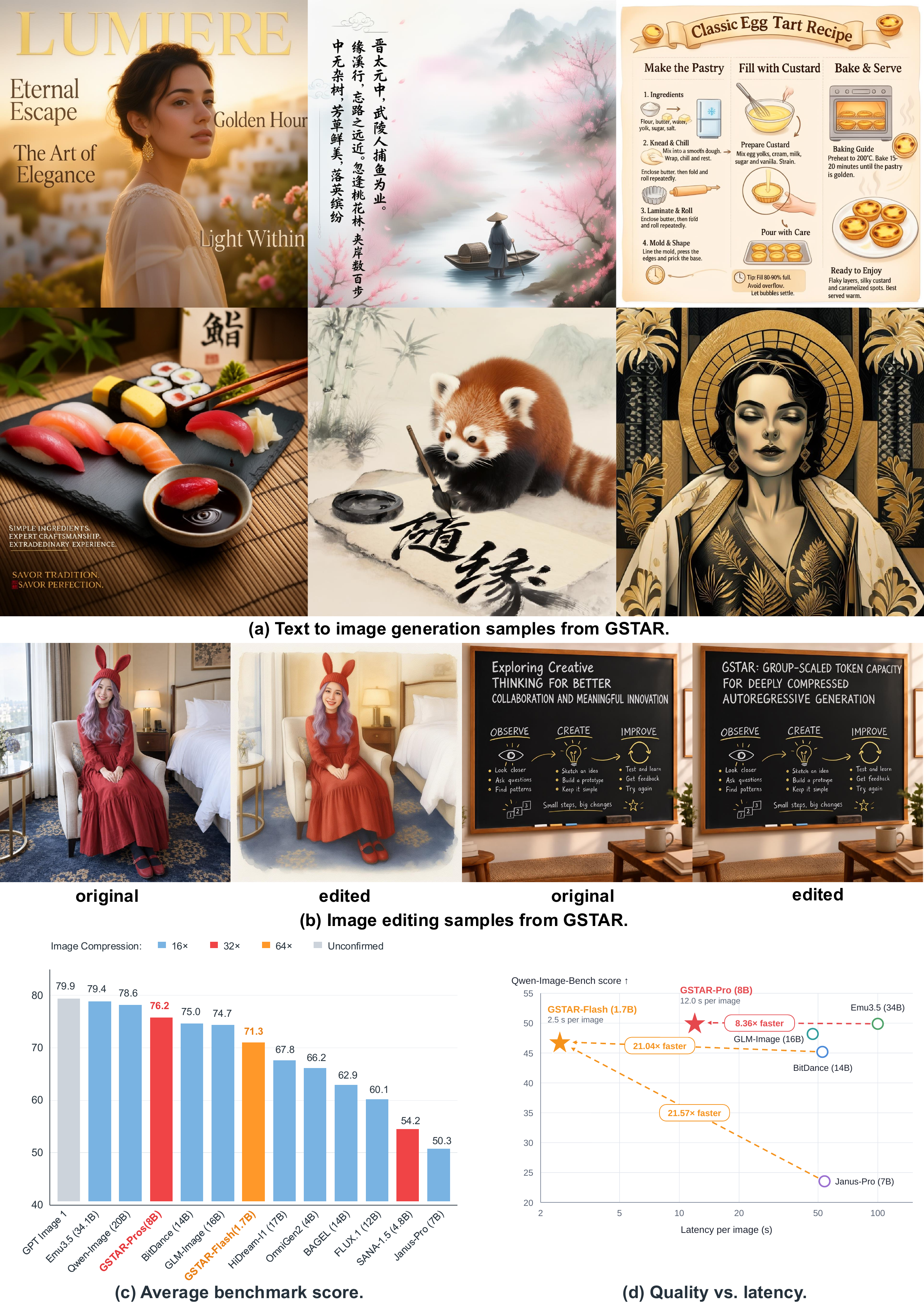}
  \caption{Overview of GSTAR\@. (a)~Text-to-image generation samples spanning portraits, landscapes, typography, and structured layouts. (b)~Image editing examples: given an original image and a text instruction, our model produces faithful edits while preserving unrelated content (original $\rightarrow$ edited). (c)~Average benchmark score comparison, where our GSTAR-Pro-8B achieves competitive performance against leading proprietary, diffusion, and discrete autoregressive models. (d)~Quality vs.\ latency at $1024\times1024$ resolution, where our GSTAR-Pro-8B and GSTAR-Flash-1.7B models achieve competitive quality with significantly lower latency.}
  \label{fig:teaser_gstar}
\end{figure*}


\section{Introduction}
\label{sec:intro}

Discrete visual tokenization provides a natural interface between image
generation and autoregressive sequence modeling. By representing images as
sequences of discrete symbols, visual generation can be formulated using the
same next-token prediction paradigm that has driven the success of large
language models \citep{esser2021taming,sun2024llamagen}. This formulation has
also shown promise for unifying visual generation and multimodal understanding
within a single autoregressive architecture
\citep{ma2025unitok,zhang2026nextflow}.

Despite its conceptual simplicity and scalability, discrete visual
autoregressive modeling faces two fundamental limitations. First, mapping
continuous visual features to a finite codebook introduces quantization error,
which can discard fine-grained details and degrade reconstruction fidelity
\citep{oord2017vqvae,esser2021taming}. The same discretization can also remove
semantically relevant cues that are important for visual understanding and
language alignment, reducing the informativeness of the resulting discrete
representation
\citep{ma2025unitok,du2025vqrae,zhang2026nextflow}. Second, discrete autoregressive models generate visual tokens sequentially,
making inference latency grow with sequence length
\citep{sun2024llamagen,wang2025par,ai2026bitdance,zhang2026nextflow}.
This bottleneck is amplified at high resolutions, where the number of tokens
grows quadratically with image side length, widening the latency gap with
diffusion models that update all spatial positions in parallel at each
denoising step \citep{ho2020ddpm_original,rombach2022ldm_original}.

Increasing the spatial compression rate offers a direct way to reduce the
number of global autoregressive steps. However, aggressive spatial compression
forces each spatial token to encode more information
\citep{chen2025dcae,chen2025dcae15}, exacerbating the information loss introduced
by discrete quantization \citep{oord2017vqvae,esser2021taming}. To mitigate this
loss, we adopt multi-codebook quantization (MCQ)
\citep{lee2022rqtransformer,xie2026videorae,ma2025unitok}, which represents each
spatial token with multiple discrete codes and thereby expands its
representation capacity. More importantly, MCQ admits a useful scaling
property: as detailed in Section~\ref{sec:group_scaling}, jointly scaling the
spatial compression rate and the number of codes per token can preserve the
total nominal information budget while reducing the number of global
autoregressive steps. We systematically study how code scaling affects
reconstruction, generation, and semantic representation, and find that it
enables high-compression tokenizers at $32\times$ and even $64\times$---well
beyond the common $16\times$ setting---while retaining high fidelity and
providing an efficient interface for downstream generation.


Multi-codebook quantization requires each global autoregressive step to predict a group of codes. RQ-Transformer-style causal depth decoders model their dependencies, but their latency grows linearly with group size and can dominate the global backbone under aggressive compression \citep{lee2022rqtransformer}. Parallel and masked-token predictors remove this serial cost \citep{chang2022maskgit,wang2025par,nie2025llada,li2026nemotron,chen2026dflash}, yet sample the codes independently and may combine incompatible modes---a failure known as multi-modal collision \citep{cheng2026dspark}. We therefore introduce a lightweight semi-autoregressive depth decoder: most layers run in parallel, while only the final layer retains a causal path for intra-group conditioning. This keeps group-scaling overhead nearly flat without sacrificing generation quality. 

We evaluate our approach on large-scale text-to-image generation using five
representative benchmarks: GenEval, DPG-Bench, PRISM, LongText-Bench, and
Qwen-Image-Bench. Our models perform competitively with leading systems across
these benchmarks while supporting spatial compression ratios from $32\times$
to $64\times$. Controlled studies further examine how group scaling affects
reconstruction fidelity, generation quality, and semantic representation, and
compare the quality--latency trade-off among causal, parallel, and
semi-autoregressive intra-group decoders. At $1024\times1024$ resolution, our
model delivers a $20\times$ speedup over prior discrete autoregressive models
and generates each image in 2.5 seconds on a single NVIDIA H200.

Our main contributions are summarized as follows:
\begin{itemize}[leftmargin=*]
    \item We systematically study how scaling the group size affects
    reconstruction, generation, and semantic understanding under $32\times$
    and $64\times$ spatial compression, showing that scaling the per-step
    capacity preserves strong representation quality with a short global
    sequence.

    \item We propose a lightweight semi-autoregressive depth decoder for
    efficiently modeling intra-group dependencies, together with an optimized
    inference implementation. Our system generates a $1024\times1024$ image in
    \textbf{2.5 seconds} on a single NVIDIA H200.

    \item We further validate the effectiveness of our approach at $32\times$ and even effective $64\times$ spatial compression by training two models, \textbf{GSTAR-Pro-8B} and \textbf{GSTAR-Flash-1.7B}, respectively, in a large-scale text-to-image setting. Our models achieve performance competitive with leading models across Qwen-Image-Bench, PRISM, DPG-Bench, GenEval, and LongText-Bench, while our approach delivers a \textbf{$20\times$} inference speedup over prior discrete autoregressive models.
\end{itemize}

\section{Related Work}
\label{sec:related_work}

\subsection{Discrete Image Generation}
Discrete generators learn distributions over quantized visual tokens
\citep{oord2017vqvae,esser2021taming,sun2024llamagen}.
Recent advances include next-scale prediction in VAR, Infinity, and NextFlow
\citep{tian2024var,han2025infinity,zhang2026nextflow}, randomized generation
orders in RAR and RandAR \citep{yu2025rar,pang2024randar}, and parallel
decoding in PAR and BitDance \citep{wang2025par,ai2026bitdance}.
Janus-Pro and Emu3.5 scale discrete generation within unified multimodal
systems \citep{chen2025januspro,cui2025emu35}; UniAR further combines a shared
visual tokenizer with parallel bitwise prediction \citep{peng2026uniar}.
GLM-Image pairs autoregressive visual-token prediction with a diffusion
image decoder \citep{zai2026glmimage}.

Representation capacity is another key axis. RQ-Transformer, MAGVIT-v2,
GSQ, UniTok, and WeTok explore residual, binary, or grouped quantization
\citep{lee2022rqtransformer,yu2024magvitv2,wang2024gsq,ma2025unitok,zhuang2025wetok},
while BAR and BitDance model large binary token spaces with masked-bit and
diffusion heads, respectively \citep{yu2026bar,ai2026bitdance}.
Building on these advances, we study spatial--group co-scaling at $32\times$
and effective $64\times$ compression, jointly examining reconstruction,
semantic representation, and the trade-off of group decoding.

\subsection{Discrete Diffusion for Accelerated Generation}
MaskGIT and Muse accelerate image generation through iterative parallel
unmasking \citep{chang2022maskgit,chang2023muse}. Show-o combines discrete
diffusion with autoregressive language modeling \citep{xie2025showo}, while
Emu3.5 adapts pretrained autoregressive models to parallel image-token
prediction via DiDA \citep{cui2025emu35}.
NVIDIA's Nemotron-Labs-Diffusion-Image advances this direction with token
editing and grouped cross-entropy for large-vocabulary image synthesis
\citep{li2026nemotron}.

For speculative language decoding, DFlash uses block-diffusion drafting
\citep{chen2026dflash}, whereas DSpark combines a parallel backbone with a
lightweight sequential module to model intra-block dependencies
\citep{cheng2026dspark}.
Inspired by these approaches, we accelerate intra-group code generation by
computing representations for all group slots in parallel, while using a
lightweight single-layer causal Transformer to model dependencies among codes.

\section{Preliminaries}
\label{sec:preliminaries}
\paragraph{Multi-codebook quantization.}
Multi-codebook quantization (MCQ) represents each spatial token with multiple
discrete codes. We instantiate MCQ with product quantization (PQ), implemented
as group-wise quantization, for its simplicity and favorable scaling: increasing
the number of groups consistently improves reconstruction fidelity
(Sec.~\ref{sec:group_scaling_exp})
\citep{jegou2011product,wang2024gsq,ma2025unitok,zhuang2025wetok}.
Specifically, PQ partitions the projected feature
$\mathbf{z}_t\in\mathbb{R}^{C}$ into $G_q$ disjoint channel groups and quantizes
each group independently:
\begin{equation}
k_t^{(g)}=\arg\min_{k\in[K]}
\left\|\mathbf{z}_t^{(g)}-\mathbf{e}_k^{(g)}\right\|_2^2,
\qquad
\hat{\mathbf{z}}_t=
\operatorname{Concat}_{g=1}^{G_q}
\mathbf{e}_{k_t^{(g)}}^{(g)}.
\end{equation}
The resulting code tuple provides up to $K^{G_q}$ configurations per spatial
position without increasing the spatial sequence length.

\paragraph{Depth transformer.}
RQ-Transformer separates the generation of residual codes into a global
spatial transformer and a local depth transformer
\citep{lee2022rqtransformer}. The global transformer summarizes preceding
spatial units into a context vector, while the depth transformer models the
dependencies among the codes assigned to the current unit. We adopt this
spatial--depth decomposition for grouped rather than residual codes; the
alternative depth-modeling strategies are detailed in
Sec.~\ref{sec:group_modeling}.

\section{Method}
\label{sec:method}

Our framework combines a deeply compressed discrete tokenizer with a hierarchical autoregressive generator. We use group scaling to increase the capacity of each spatial token and develop a spatial--group co-scaling principle that preserves the nominal information budget while shortening the global sequence. Building on the native $32\times$ tokenizer, we instantiate this principle through lossless spatial packing to construct an effective $64\times$ tokenizer. A lightweight semi-autoregressive depth decoder then efficiently models the correlated codes within each global step.

\begin{figure*}[!t] \centering \vspace{2pt} \includegraphics[width=\textwidth]{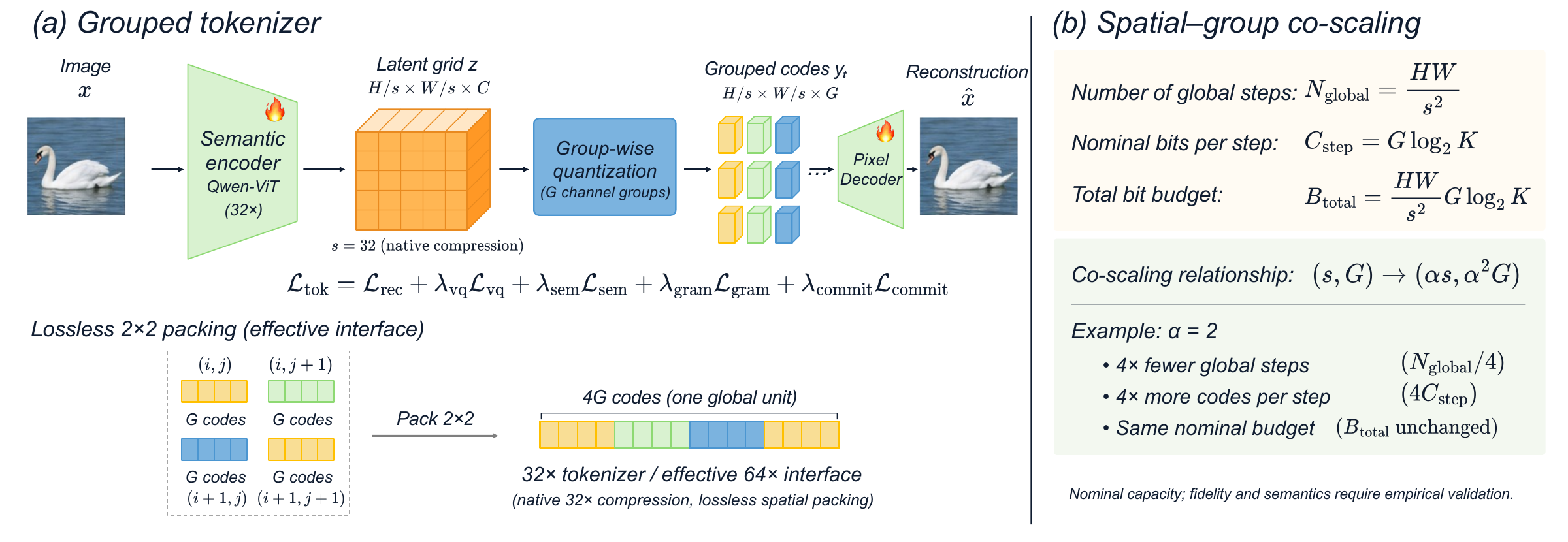} \caption{Grouped tokenization and spatial–group co-scaling. (a) A semantic encoder maps an
image to a spatially compressed latent grid, and group-wise quantization represents each position
with G discrete codes. Lossless 2 × 2 packing converts the native 32× grid into an effective 64×
interface, reducing the number of global autoregressive steps by four while increasing the codes per
step from G to 4G. (b) Jointly scaling spatial compression and group size preserves the nominal
information capacity per image while shortening the global sequence.} 
\vspace{-10pt}
\label{fig:draft_tokenizer_scaling} \end{figure*}

\subsection{Deeply Compressed Tokenizer for Efficient Generation}
\label{sec:deep_tokenizer}

We train a discrete tokenizer that maps an image $\mathbf{x}$ to a compact grid of group-wise codes and reconstructs it as $\hat{\mathbf{x}}$, as illustrated in Fig.~\ref{fig:draft_tokenizer_scaling}(a). Following recent visual autoencoders and tokenizers built on pretrained semantic encoders \citep{yao2025vavae,du2025vqrae,ma2026drae,zheng2026rae,huang2026dcsaedeepcompressionsemantic}, we initialize the encoder of our $32\times$ tokenizer from Qwen-ViT \citep{bai2025qwen3vl}. The tokenizer is optimized with standard reconstruction and vector-quantization objectives. To preserve the semantic structure of the pretrained representation after quantization, we additionally align the quantized features with the corresponding Qwen-ViT features. Following Scale-RAE for text-to-image generation \citep{tong2026scalerae}, we further employ a Gram loss \citep{gatys2015neural} to match second-order feature correlations between the input and its reconstruction. The overall training objective is
\begin{equation}
    \mathcal{L}_{\mathrm{tok}}
    = \mathcal{L}_{\mathrm{rec}}
    + \lambda_{\mathrm{vq}}\mathcal{L}_{\mathrm{vq}}
    + \lambda_{\mathrm{sem}}\mathcal{L}_{\mathrm{sem}}
    + \lambda_{\mathrm{gram}}\mathcal{L}_{\mathrm{gram}} + \lambda_{\mathrm{commit}}\mathcal{L}_{\mathrm{commit}}.
\end{equation}
Here $\mathcal{L}_{\mathrm{rec}}$ collects the standard pixel, perceptual, and adversarial reconstruction terms, $\mathcal{L}_{\mathrm{vq}}$ is the quantization objective, and $\mathcal{L}_{\mathrm{sem}}$ is the semantic alignment objective. The commitment loss $\mathcal{L}_{\mathrm{commit}}$ encourages encoder features to remain close to their assigned codebook embeddings.

\subsection{Scaling the Group Size}
\label{sec:group_scaling}

Group-wise quantization provides a direct way to increase the discrete capacity assigned to each spatial token without lengthening the global sequence. This is particularly useful under aggressive spatial compression, where each remaining position must preserve more visual information. Let's denote the effective per-axis spatial compression factor and $G$ the number of codes generated at each global position. For the native $32\times$ tokenizer, $G=G_q$, the number of quantization groups. With a fixed codebook size $K$, the relationships illustrated in Fig.~\ref{fig:draft_tokenizer_scaling}(b) are
\begin{equation}
\begin{aligned}
N_{\mathrm{global}}(s) &= \tfrac{HW}{s^2},\quad
    C_{\mathrm{step}}(G,K) = G\log_2 K, \\
B_{\mathrm{total}}(s,G,K)
    &= N_{\mathrm{global}}\,C_{\mathrm{step}}
     = \tfrac{HW}{s^2}\,G\log_2 K.
\end{aligned}
\label{eq:global_capacity}
\end{equation}
Here $B_{\mathrm{total}}$ is a fixed-length nominal bit budget rather than the empirical entropy or effective semantic information of the learned codes.

The two variables control complementary properties. At fixed $s$, increasing $G$ enlarges the representation capacity without increasing the number of global autoregressive steps. Increasing $s$ 
shortens the global sequence quadratically. These properties motivate spatial--group co-scaling: scaling $s$ by $\alpha$ and $G$ by $\alpha^2$ preserves $B_{\mathrm{total}}$ while reducing the number of global steps by a factor of $\alpha^2$.

\paragraph{Constructing the effective $64\times$ tokenizer.}
Guided by this relation, we construct an effective $64\times$ tokenizer from the native $32\times$ tokenizer through the $2\times2$ spatial packing shown in Fig.~\ref{fig:draft_tokenizer_scaling}(a). Each neighborhood of four quantized positions is packed into one global unit, increasing the codes per step from $G_q$ to $4G_q$ while shortening the global sequence by a factor of four. This realizes the co-scaling relation with $\alpha=2$, preserving all quantized codes without introducing additional quantization. The packed codes are restored to the native grid before image decoding.

To validate this capacity analysis, we systematically vary the group size and evaluate reconstruction fidelity, semantic representation, and downstream generation in Sec.~\ref{sec:group_scaling_exp}. The results confirm that increasing group size provides additional usable information capacity and improves representation quality. Larger code tuples, however, also increase the cost of intra-group dependency modeling, motivating the efficient depth decoder introduced next.

\begin{figure*}[!t] \centering \vspace{2pt} \includegraphics[width=\textwidth]{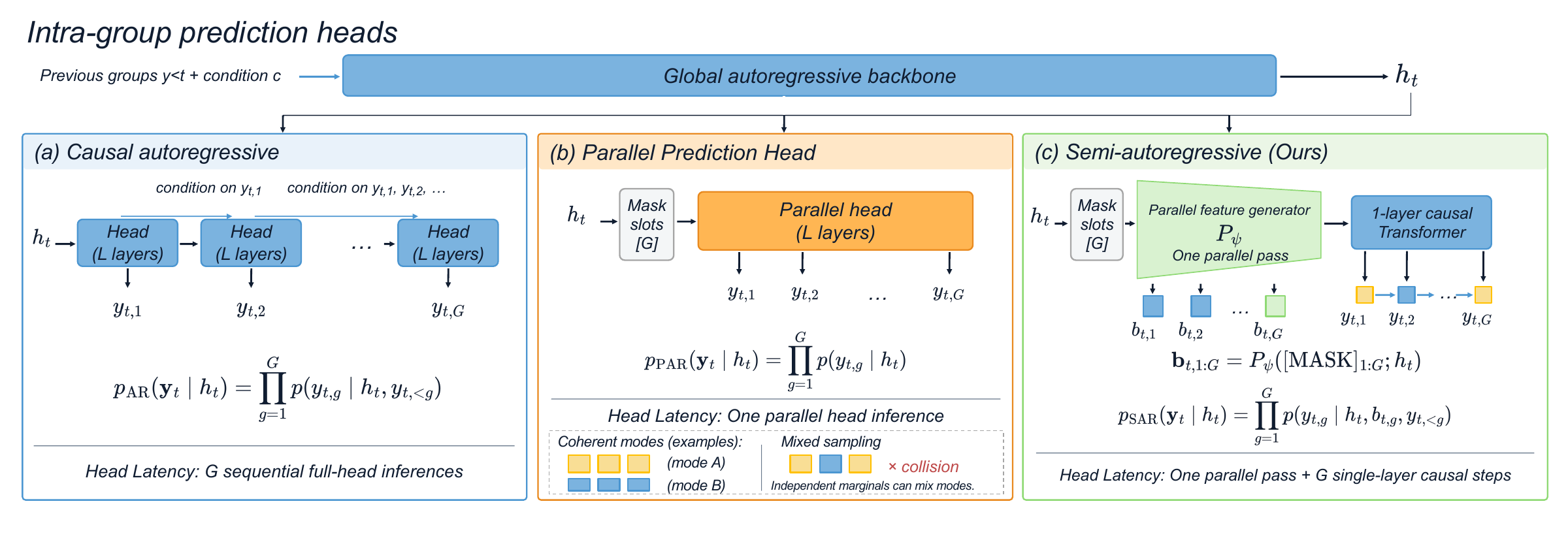} \caption{Comparison of intra-group prediction heads. Given the backbone context ht, (a) causal
autoregressive decoding captures dependencies among codes but requires repeated, expensive head
evaluations, while (b) parallel prediction head is fast but ignores these dependencies, risking
inconsistent code combinations. (c) Our semi-autoregressive head addresses both limitations: a par-
allel feature generator performs the heavy computation once, and a lightweight single-layer causal
Transformer models dependencies during sequential sampling.} 
\label{fig:draft_intra_group_heads} \end{figure*}

\subsection{Efficient Intra-Group Modeling}
\label{sec:group_modeling}

Group scaling requires each global autoregressive step to generate a tuple of
correlated discrete codes. A common approach, following RQ-Transformer
\citep{lee2022rqtransformer}, is to use a causal depth head to model intra-group
dependencies, but its sequential decoding incurs increasing inference latency
as the group size grows. Fully parallel prediction removes this serial
overhead, yet sacrifices generation quality by ignoring these dependencies.
To reconcile these competing objectives, we propose a semi-autoregressive head
that preserves causal dependency modeling while approaching the decoding speed
of parallel prediction. Below, we first examine the causal and parallel
alternatives, then present our design.

At global decoding step $t$, the autoregressive backbone produces a context
representation $\mathbf{h}_t$. The intra-group head uses this context to generate
$G$ codes, $\mathbf{y}_t=(y_{t,1},\ldots,y_{t,G})$, by modeling the conditional
joint distribution $p(\mathbf{y}_t\mid\mathbf{h}_t)$.

\paragraph{Causal Autoregressive Head.}
Following the spatial--depth decomposition of RQ-Transformer and its use in
grouped-code generation \citep{lee2022rqtransformer,ma2025unitok}, a causal
autoregressive head predicts the codes one by one, conditioning each prediction
on the global context and previously generated codes:
\vspace{-2pt}
\begin{equation}
    p_{\mathrm{AR}}(\mathbf{y}_t\mid\mathbf{h}_t)
    =
    \prod_{g=1}^{G}
    p(y_{t,g}\mid\mathbf{h}_t,\mathbf{y}_{t,<g}).
\end{equation}
This factorization preserves intra-group dependencies, but requires running
the full head for each of the $G$ sequential predictions. As $G$ grows, this
serial computation becomes a substantial decoding bottleneck and can dominate
the global backbone under aggressive compression.

\paragraph{Parallel Prediction Head.}
To remove this sequential overhead, a fully parallel head predicts all $G$
codes in a single forward pass:
\begin{equation}
    p_{\mathrm{PAR}}(\mathbf{y}_t\mid\mathbf{h}_t)
    =
    \prod_{g=1}^{G}
    p(y_{t,g}\mid\mathbf{h}_t).
\end{equation}
This removes the intra-group serial path
\citep{chang2022maskgit,wang2025par,nie2025llada,li2026nemotron}, but makes the
codes conditionally independent during sampling. Since codes within a group
jointly describe the same visual content, independently sampled marginals may
select incompatible modes, leading to \emph{multimodal collision}
\citep{cheng2026dspark}. The quality degradation motivates a head
that retains causal conditioning while keeping most computation parallel.

\paragraph{Semi-Autoregressive Head.}
We propose a semi-autoregressive head that combines a parallel trunk with a
lightweight causal module, retaining intra-group dependencies without repeating
expensive representation modeling $G$ times. Drawing architectural inspiration
from the parallel--sequential drafters in DFlash and DSpark
\citep{chen2026dflash,cheng2026dspark}, the parallel trunk first computes
representations for all group slots in one pass:
\begin{equation}
    \mathbf{b}_{t,1:G}
    = P_{\psi}([\textsc{Mask}]_{1:G};\mathbf{h}_t).
\end{equation}
A lightweight, single-layer causal Transformer then generates each code conditioned on
both its parallel representation and the previously sampled intra-group codes:
\begin{equation}
    p_{\mathrm{SAR}}(\mathbf{y}_t\mid\mathbf{h}_t)
    =
    \prod_{g=1}^{G}
    p(y_{t,g}\mid
      \mathbf{h}_t,\mathbf{b}_{t,g},\mathbf{y}_{t,<g}).
\end{equation}
The parallel trunk is evaluated only once per global position. Only the
single-layer, KV-cached causal Transformer remains sequential, allowing each
code to depend on previously sampled codes without repeatedly executing the
full head. Experiments in Sec.~\ref{sec:head_ablation} show that our head
achieves generation quality comparable to the fully causal head with latency
close to parallel prediction, while consistently outperforming the parallel
head across all three quality metrics.

\section{Experiments}
\label{sec:experiments}

\paragraph{Experimental setup.}
We develop two models: \textbf{GSTAR-Pro-8B}, which uses a $32\times$ tokenizer with group size $G=16$, and \textbf{GSTAR-Flash-1.7B}, which uses an effective $64\times$ tokenizer with $G=32$ for fast image generation. Following prior progressive-training recipes~\citep{deng2025bagel,chen2025sanavideo,chen2026sanavideo2}, both models are trained through pretraining, continued training (CT), and supervised fine-tuning (SFT), with progressively increasing image resolution.Detailed tokenizer training procedures and stage-wise generator training configurations are provided in Appendix~\ref{app:training_details}. Both models support image generation and editing, evaluated in Sec.~\ref{sec:main_experiments}. We further analyze the effects of group scaling on reconstruction, generation, and semantic understanding in Sec.~\ref{sec:group_scaling_exp}, compare the quality and latency of different prediction heads in Sec.~\ref{sec:head_ablation}, and investigate the sources of acceleration in Sec.~\ref{sec:efficiency}.

\subsection{Main Results}
\label{sec:main_experiments}
We adopt GenEval~\citep{ghosh2023geneval}, DPG-Bench~\citep{hu2024ella}, PRISM~\citep{fang2026prism}, LongText-Bench~\citep{geng2025xomni}, and Qwen-Image-Bench~\citep{li2026qwenimagebench} to evaluate text-to-image generation, covering compositional alignment, perceptual quality, text rendering, and broader prompt-following capabilities. We evaluate instruction-guided image editing on GEdit-Bench~\citep{liu2025step1xedit}, reporting semantic consistency, perceptual quality, and overall scores for its English subset. Results for GSTAR-Pro-8B and GSTAR-Flash-1.7B are reported in Tables~\ref{tab:t2i_main} and~\ref{tab:editing_main}, respectively for generation and editing. Benchmark versions, sampling configurations, and evaluation protocols are detailed in Appendix~\ref{app:evaluation_details}. We also evaluate the reconstruction quality of our visual tokenizers in Appendix~\ref{app:reconstruction}.


\begin{table}[t]
\centering
\small
\setlength{\tabcolsep}{3.5pt}
\renewcommand{\arraystretch}{1.03}
\caption{
Text-to-image generation results.
PRISM is the English overall combined score
(Qwen2.5-VL-72B judge), rescaled from 0--100 to 0--10.
LongText reports the English subset; Qwen-Image denotes
Qwen-Image-Bench overall (0--100).
}
\label{tab:t2i_main}
\resizebox{\linewidth}{!}{%
\begin{tabular}{@{}lccrrrrr@{}}
\toprule
Model & Comp. & Params.
& GenEval $\uparrow$
& DPG $\uparrow$
& PRISM $\uparrow$
& LongText $\uparrow$
& Qwen-Image $\uparrow$ \\
\midrule

\multicolumn{8}{@{}l}{\textit{Proprietary Models}} \\
Kolors 2.0
& -- & --
& -- & -- & -- & 0.258 & -- \\
GPT Image 1
& -- & --
& 0.84 & 85.15 & 8.07 & 0.956 & 54.07 \\
Seedream 3.0
& -- & --
& 0.84 & 88.27 & 7.62 & 0.896 & -- \\

\midrule
\multicolumn{8}{@{}l}{\textit{Diffusion Models}} \\
HiDream-I1-Full
& $16\times$ & 17B
& 0.83 & 85.89 & 7.26 & 0.543 & 43.08 \\
FLUX.1 [dev]
& $16\times$ & 12B
& 0.66 & 83.84 & 6.85 & 0.607 & 21.58 \\
BAGEL
& $16\times$ & 14B
& 0.88 & 85.07 & 6.20 & 0.373 & 42.04 \\
Z-Image
& $16\times$ & 6B
& 0.84 & 88.14 & 7.56 & 0.935 & 49.10 \\
Qwen-Image
& $16\times$ & 20B
& 0.87 & 88.32 & 7.41 & 0.943 & 49.23 \\
OmniGen2
& $16\times$ & 4B
& 0.86 & 83.57 & 6.48 & 0.561 & 40.70 \\
SANA-1.5
& $32\times$ & 4.8B
& 0.81 & 84.70 & 6.51 & 0.050 & 35.14 \\

\midrule
\multicolumn{8}{@{}l}{\textit{Autoregressive Models}} \\
Janus-Pro-7B
& $16\times$ & 7B
& 0.80 & 84.19 & 6.21 & 0.019 & 23.53 \\
GLM-Image
& $16\times$ & 16B
& 0.72 & 84.78 & 7.32 & 0.952 & 48.19 \\
Emu3.5
& $16\times$ & 34.1B
& 0.86 & 88.26 & 7.52 & 0.976 & 49.89 \\
BitDance
& $16\times$ & 14B
& 0.86 & 88.28 & 7.19 & 0.837 & 45.19 \\

\midrule
\textbf{(Ours)GSTAR-Pro-8B}
& $\mathbf{32\times}$ & \textbf{8B}
& 0.84 & 83.38 & 6.98 & 0.935 & 50.07 \\
\textbf{(Ours)GSTAR-Flash-1.7B}
& $\mathbf{64\times}$ & \textbf{1.7B}
& 0.76
& 80.18
& 6.81
& 0.853
& 46.88 \\
\bottomrule
\end{tabular}%
}
\end{table}

\subsection{Ablation Studies}
\label{sec:ablations}

\subsubsection{Group Scaling}
\label{sec:group_scaling_exp}%
\begin{wraptable}{r}{0.48\textwidth}
\vspace{-14pt}
\centering\footnotesize\setlength{\tabcolsep}{3pt}
\caption{Effects of group size with $K=16{,}384$.}
\label{tab:group_scaling}
\begin{tabular}{@{}crrr@{}}
\toprule
$G$ & rFID $\downarrow$ & gFID $\downarrow$
& \shortstack{Linear probe $\uparrow$\\(\%)} \\
\midrule
1  & 2.0624 & \textbf{2.1057} & 81.45 \\
8  & 0.8525 & 6.3773        & 82.90 \\
16 & 0.5357 & 5.7388          & 84.15 \\
32 & \textbf{0.4325} & 17.05  & \textbf{84.55} \\
\bottomrule
\end{tabular}
\vspace{-8pt}
\end{wraptable}
We vary the group size $G\in\{1,8,16,32\}$ while fixing the
codebook size at $K=16{,}384$ and the spatial compression factor at
$32\times$, and examine how increased per-position capacity affects
reconstruction, generative modeling, and semantic understanding.
Training details for the ImageNet ablations are provided in
Appendix~\ref{app:imagenet_ablation_training}. Generally, increasing $G$ improves reconstruction and semantic
representation (Table~\ref{tab:group_scaling}), but slows early-stage
generation convergence, consistent with
VA-VAE~\citep{yao2025vavae}. Suprisingly, $G=16$ eventually catches
up with and outperforms $G=8$, despite converging more slowly at first
(Fig.~\ref{fig:imagenet_group_scaling}).

\begin{wraptable}{r}{0.46\textwidth}
\vspace{-14pt}
\centering\footnotesize\setlength{\tabcolsep}{3pt}
\caption{Text-to-image group scaling. PRISM: combined score (0--10).}
\label{tab:group_scaling_t2i}
\begin{tabular}{@{}crrr@{}}
\toprule
$G$ & GenEval $\uparrow$ & DPG $\uparrow$
& PRISM $\uparrow$ \\
\midrule
8  & 0.8060 & 82.49 & 5.30 \\
16 & 0.8386 & 82.48 & 5.46 \\
\bottomrule
\end{tabular}
\vspace{-12pt}
\end{wraptable}
We further validate this observation on text-to-image
generation using the same setting as the head ablation
in Sec.~\ref{sec:head_ablation}: Qwen3-1.7B initialization,
a $32\times$ tokenizer, 100k pretraining steps at
$512\times512$ resolution, a sequence length of 8,192,
and a global batch size of 32, followed by one epoch of SFT.
Table~\ref{tab:group_scaling_t2i} shows that $G=16$
achieves comparable or slightly better performance.

Together with the late-stage ImageNet results, these findings suggest that
larger groups provide a richer representational space that preserves more
visual and semantic information. Although the resulting space is more
difficult to model and leads to slower early-stage convergence, sufficient
training allows the generator to exploit its increased capacity. Larger
models may benefit even further from this additional capacity.

We also validate that our lossless packing strategy, which transforms
$\mathrm{f32g8}$ into $\mathrm{f64g32}$, retains strong text-to-image
generation performance; see Appendix~\ref{app:lossless_packing}.

\subsubsection{Intra-Group Prediction Heads}
\label{sec:head_ablation}

We compare the Causal AR Head, Parallel Prediction Head,
and our Semi-AR Head using a Qwen3-1.7B-initialized model
with a $32\times$ tokenizer and $G=16$.
All variants share the same training setting: 100k pretraining
steps at a fixed $512\times512$ resolution, a sequence length
of 8,192, and a global batch size of 32, followed by one
epoch of SFT.

As shown in Table~\ref{tab:head_ablation}, our Semi-AR Head
substantially reduces inference latency while maintaining
generation quality comparable to the Causal AR Head.
It also consistently outperforms the Parallel Prediction
Head across all three quality metrics.

\begin{table}[t]
\centering
\setlength{\abovecaptionskip}{2pt}
\setlength{\belowcaptionskip}{1pt}
\begin{minipage}[t]{0.453\linewidth}
\vspace{0pt}

\centering
\includegraphics[width=\linewidth]{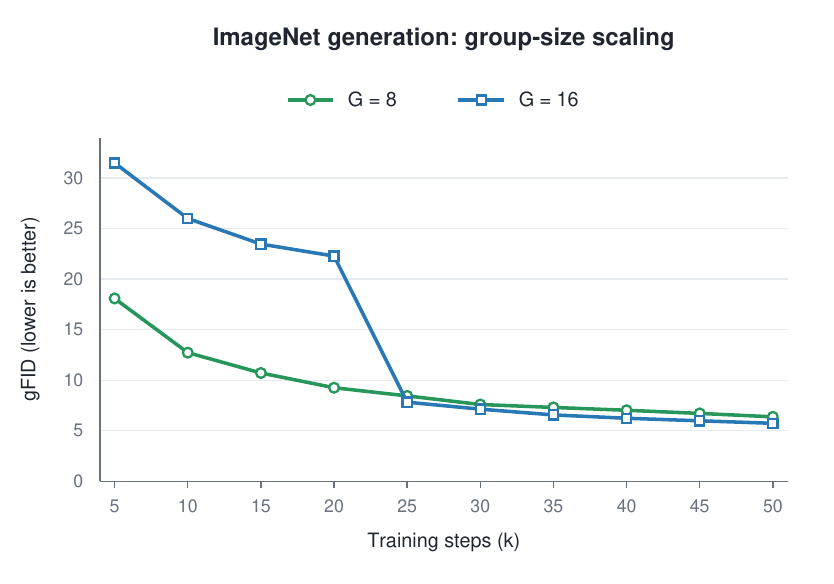}
\captionof{figure}{ImageNet gFID during training. $G=16$ converges more slowly at first but outperforms $G=8$ from 25k steps.}
\label{fig:imagenet_group_scaling}

\vspace{2pt}
\centering
\footnotesize
\setlength{\tabcolsep}{2.5pt}
\captionof{table}{Prediction heads with matched training. PRISM: combined score; latency is measured per image at $512\times512$.}
\label{tab:head_ablation}
\resizebox{\linewidth}{!}{%
\begin{tabular}{@{}lrrrr@{}}
\toprule
Head & \shortstack{GenEval\\$\uparrow$} & \shortstack{DPG\\$\uparrow$}
& \shortstack{PRISM\\$\uparrow$} & \shortstack{Head $\downarrow$\\(s)} \\
\midrule
Causal AR
& \textbf{0.8386} & 82.48 & \textbf{5.46}
& 6.44 \\
Parallel
& 0.7800 & 79.98 & 4.72
& \textbf{1.77} \\
Semi-AR (ours)
& 0.8253 & \textbf{82.60} & \textbf{5.46}
& 3.41 \\
\bottomrule
\end{tabular}}

\end{minipage}\hfill
\begin{minipage}[t]{0.52\linewidth}
\vspace{0pt}
\centering\footnotesize\setlength{\tabcolsep}{3pt}
\caption{Image editing on GEdit-Bench (English full set).}
\label{tab:editing_main}
\resizebox{\linewidth}{!}{%
\begin{tabular}{@{}lrrr@{}}\toprule
Model & SC $\uparrow$ & PQ $\uparrow$ & Overall $\uparrow$ \\\midrule
\multicolumn{4}{@{}l}{\textit{Proprietary Models}} \\
Gemini 2.0 & 6.73 & 6.61 & 6.32 \\
Gemini 2.5 Flash Image & 7.41 & 7.96 & 7.10 \\
GPT-Image-1 [High] & 7.85 & 7.62 & 7.53 \\
\midrule
\multicolumn{4}{@{}l}{\textit{Open Models}} \\
AnyEdit & 3.18 & 5.82 & 3.21 \\
Instruct-Pix2Pix & 3.58 & 5.49 & 3.68 \\
MagicBrush & 4.68 & 5.66 & 4.52 \\
UniWorld-v1 & 4.93 & 7.43 & 4.85 \\
OmniGen & 5.96 & 5.89 & 5.06 \\
FLUX.1 Kontext [Dev] & 6.52 & 7.38 & 6.00 \\
OmniGen2 & 7.16 & 6.77 & 6.41 \\
BAGEL & 7.36 & 6.83 & 6.52 \\
Step1X-Edit & 7.66 & 7.35 & 6.97 \\
Qwen-Image-Edit & 8.00 & 7.86 & 7.56 \\
Emu3.5 & 8.11 & 7.70 & 7.59 \\
\midrule
GSTAR-Pro-8B & 7.85 & 7.45 & 7.45 \\
\bottomrule
\end{tabular}}
\vspace{-32pt}
\end{minipage}
\vspace{-14pt}
\end{table}

\subsection{Efficiency Analysis}
\label{sec:efficiency}

\begin{figure}[htbp]
\centering
\includegraphics[width=\linewidth]{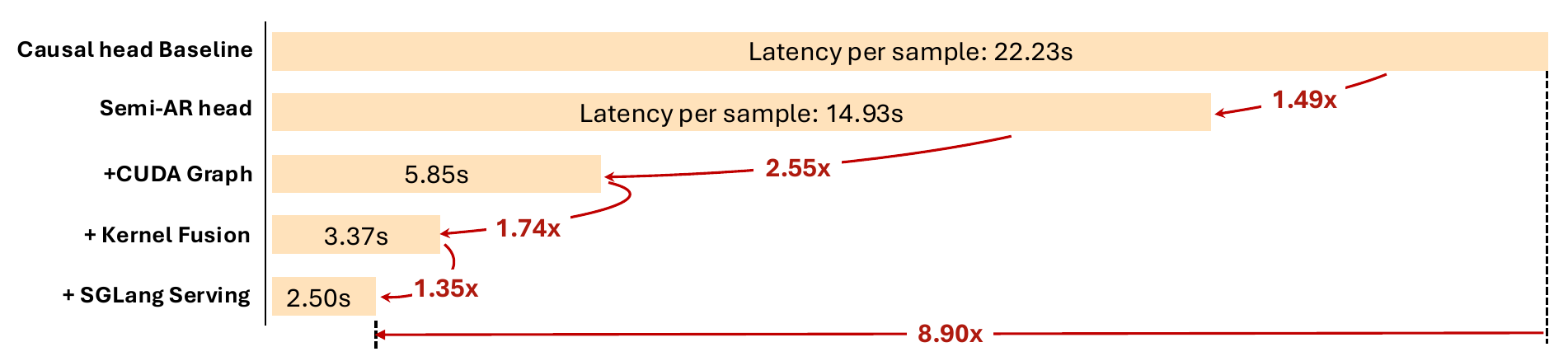}
\caption{Cumulative speedup breakdown for GSTAR-Flash-1.7B at $1024\times1024$. Starting from the Causal AR head baseline (22.23\,s), each stage---Semi-AR head, CUDA graph capture, fused top-$k$ kernel, and SGLang serving---contributes multiplicative savings, reaching $8.9\times$ end-to-end.}
\label{fig:speedup}
\end{figure}

Our framework enables highly compressed visual representations
that shorten the global autoregressive sequence, while the
Semi-AR Head reduces the latency of intra-group modeling.
Together, these designs reduce decoding costs at both the
global and local levels.
Beyond the architectural savings, we further optimize the system-level inference pipeline. We measure per-image latency for GSTAR-Flash-1.7B generating at $1024\times1024$ and apply four cumulative optimizations (Figure~\ref{fig:speedup}).
Starting from a Causal AR head baseline of 22.23\,s, replacing it with our Semi-AR Head reduces latency to 14.93\,s ($1.49\times$) by parallelizing intra-group token prediction.
The remaining optimizations target the two main bottlenecks separately: the intra-group head and the LLM backbone.
For the head, we first capture each global step as a single CUDA graph, eliminating kernel-dispatch overhead ($2.55\times$, 5.85\,s).
We then fuse chains of small elementwise and reduction kernels in the sampling and normalization paths into composite Triton kernels, reducing the per-block kernel count and eliminating redundant global-memory round trips ($1.74\times$, 3.37\,s).
Finally, integrating the model into SGLang~\citep{zheng2024sglang} for LLM backbone serving yields a further $1.35\times$ speedup (2.50\,s).
The four stages compound to an $8.9\times$ end-to-end speedup.

\section{Conclusion}
We explore high-quality autoregressive image generation under deep
spatial compression by scaling discrete token capacity.
Combining grouped quantization with efficient semi-autoregressive
intra-group modeling enables generation at $32\times$ and $64\times$
spatial downsampling.
Our GSTAR-Pro-8B and GSTAR-Flash-1.7B models demonstrate the effectiveness of
this approach in large-scale text-to-image generation, achieving
performance competitive with leading models and a $20\times$
speedup over prior discrete autoregressive models.


\bibliographystyle{plainnat}
\setlength{\bibhang}{0pt}
\setlength{\bibindent}{0pt}
\bibliography{iclr2027_conference}

\clearpage

\appendix
\section{Training Details}
\label{app:training_details}

\subsection{Model Training}
\label{app:model_training}

\paragraph{Initialization and optimization.}
We initialize GSTAR-Pro-8B from Qwen3-8B and GSTAR-Flash-1.7B from Qwen3-1.7B.
Both models share the same three-stage training schedule, consisting
of pretraining (PT), continued training (CT), and supervised
fine-tuning (SFT). We use ZeRO Stage 2 throughout all three stages
and optimize both models using Adam with $\beta_1=0.9$ and
$\beta_2=0.95$.

\paragraph{Training schedule.}
We use any-resolution training throughout all three stages,
progressively increasing the image resolution.
PT runs for 400k steps at 512px, followed by 100k CT steps at 1024px.
Both stages use a sequence length of 4,096, a global batch size
of 64, and a learning rate of $10^{-4}$.
SFT runs for 30k steps, mixing 1024px and 2048px images at a ratio
of 1:1, with a sequence length of 8,192, a global batch size of 32,
and a learning rate of $10^{-5}$.
During SFT, image-editing and text-to-image generation data are
mixed at a ratio of 3:7.
Table~\ref{tab:training_schedule} summarizes the configurations
shared by both models.

\begin{table}[htbp]
\centering
\small
\caption{Training configurations shared by GSTAR-Pro-8B and GSTAR-Flash-1.7B.
All stages use any-resolution training. SFT mixes the 1024px
and 2048px resolution tiers at a ratio of 1:1.}
\label{tab:training_schedule}
\begin{tabular}{lccccc}
\toprule
Stage & Resolution & Steps & Seq. length & Global batch & LR \\
\midrule
PT  & 512px            & 400k & 4,096 & 64 & $10^{-4}$ \\
CT  & 1024px           & 100k & 4,096 & 64 & $10^{-4}$ \\
SFT & 1024px / 2048px  & 30k  & 8,192 & 32 & $10^{-5}$ \\
\bottomrule
\end{tabular}
\end{table}

\paragraph{Training data.}
For PT and CT, we collect image--text pairs from publicly available
datasets, including LAION, DataComp, and JourneyDB, and regenerate
their captions using Qwen3.5-122B-A10B.
For SFT, we select a higher-quality subset of these data and
supplement it with a small proportion of images regenerated using
Seedream and Nano Banana 2.

\subsection{Tokenizer Training}
\label{app:tokenizer_training}

\paragraph{Initialization and objectives.}
We initialize the tokenizer encoder from Qwen-ViT with a spatial
downsampling factor of $32$. For both the ImageNet ablations and
text-to-image tokenizers, we use the same loss weights:
$\lambda_{\mathrm{sem}}=1$, $\lambda_{\mathrm{gram}}=10$,
$\lambda_{\mathrm{perceptual}}=1$, $\lambda_{\mathrm{vq}}=1$,
and $\lambda_{\mathrm{commit}}=0.25$.
Here $\lambda_{\mathrm{perceptual}}$ weights the perceptual reconstruction
loss, and $\lambda_{\mathrm{commit}}$ weights the commitment term within
the vector-quantization objective.

\paragraph{Optimization.}
All tokenizers are trained with Adam using
$(\beta_1,\beta_2)=(0.9,0.95)$ and a global batch size of 1,024.
We use a base learning rate of $10^{-4}$ for the quantizer, decoder,
and other components. To preserve the pretrained semantic encoder's
semantic representations, we use a lower learning rate of $10^{-5}$
for the encoder.

\paragraph{Tokenizer configurations.}
For GSTAR-Pro-8B, we train a $32\times$ tokenizer with 16 quantization
groups per spatial position.
For GSTAR-Flash-1.7B, we train a $32\times$ tokenizer with 8 quantization
groups per spatial position, then merge each $2\times2$ neighborhood
of quantized positions to obtain an effective $64\times$ tokenizer
with 32 groups per global position.
This merging operation reorganizes the quantized representation
without introducing an additional quantization step.

\paragraph{Tokenizers for text-to-image generation.}
We use any-resolution training for the text-to-image tokenizers.
In addition to general-purpose image data, we incorporate OCR,
aesthetic, and face data to improve generalization and text-rendering
capabilities. Training runs for a total of 200k steps. 

Following UniAR~\citep{peng2026uniar} and Emu3.5~\citep{cui2025emu35},
we freeze the tokenizer after training and train a DiT-based visual
decoder initialized from Ideogram 4~\citep{ideogram2026}, conditioned on the discrete
tokens produced by the frozen tokenizer.
This decoder is used solely to refine visual quality.
Unlike GLM-Image~\citep{zai2026glmimage} and X-Omni~\citep{geng2025xomni},
our decoder conditions exclusively on our discrete visual tokens,
without any additional textual conditioning.
The ImageNet ablation settings are described separately in
Appendix~\ref{app:imagenet_ablation_training}.


\subsection{ImageNet Ablation Settings}
\label{app:imagenet_ablation_training}

\paragraph{Tokenizer training.}
For the ImageNet ablations, we train the tokenizers on ImageNet for
100k steps at a fixed resolution of $256\times256$.
The loss weights, optimizer, component-wise learning rates, and global
batch size follow Appendix~\ref{app:tokenizer_training}.

\paragraph{Class-conditional generation.}
For the class-to-image (C2I) ablations, we train the autoregressive
generator on ImageNet at $256\times256$ resolution using the
Llama-XL architecture from LlamaGen~\citep{sun2024llamagen}.
We train for 50k steps with a sequence length of 4,096, a global
batch size of 128, and a learning rate of $4\times10^{-4}$.
All remaining hyperparameters follow the Llama-XL configuration.
\section{Multimodal Understanding}
\label{app:multimodal_understanding}

\paragraph{Experimental setup.}
We initialize the language model from Qwen3-1.7B
and train on the standard LLaVA-OneVision training
data~\citep{li2024llavaonevision}.
We compare continuous visual representations from the original
Qwen3 ViT (without further training) with discrete representations
produced by our f32g16 tokenizer, which uses a spatial
downsampling factor of $32$ and $16$ quantization groups per position.

\paragraph{Results.}
As shown in Table~\ref{tab:multimodal_understanding}, our discrete
tokenizer achieves multimodal understanding performance comparable
to continuous visual representations, with average scores of
64.7 and 64.6, respectively.
These results suggest that our tokenizer largely preserves
the semantic information needed for multimodal understanding
despite discretization and spatial compression.

\begin{table}[htbp]
\centering
\small
\caption{Multimodal understanding performance using continuous and
discrete visual representations. Higher is better for all metrics.
The average is computed over the eight listed benchmarks, with
OCRBench divided by 10 to match the 0--100 scale of the other scores.}
\label{tab:multimodal_understanding}
\begin{tabular}{lrr}
\toprule
Benchmark & Continuous & Discrete (f32g16) \\
\midrule
MathVista~\citep{lu2023mathvista}
    & 45.7 & 44.5 \\
RealWorldQA~\citep{xai2024realworldqa}
    & 60.1 & 59.3 \\
MMStar~\citep{chen2024mmstar}
    & 45.8 & 44.7 \\
OCRBench~\citep{liu2023ocrbench}
    & 676 & 665 \\
TextVQA~\citep{singh2019textvqa}
    & 71.3 & 71.0 \\
AI2D (NO\_MASK)~\citep{kembhavi2016ai2d}
    & 76.7 & 78.6 \\
ChartQA~\citep{masry2022chartqa}
    & 67.4 & 69.9 \\
DocVQA~\citep{mathew2021docvqa}
    & 81.9 & 82.9 \\
\midrule
Average & 64.6 & 64.7 \\
\bottomrule
\end{tabular}
\end{table}

\section{Evaluation Protocols and Additional Results}
\label{app:evaluation_details}

\subsection{Latency Measurement}
\label{app:latency_protocol}
We measure batch-one generation at $1024\times1024$ on an NVIDIA H200 and separately report global-backbone time, total intra-group head time, image-decoder time, and end-to-end pipeline latency. All models run in BF16 precision. The software stack consists of PyTorch~2.9.1 with CUDA~12.8, Triton~3.5.1, and Transformers~4.57.6. The decode loop is captured as a single CUDA graph. Each configuration is warmed up for 3 iterations, then timed over 3 independent trials; we report the median latency. Sampling uses classifier-free guidance with a scale of 1.8, temperature 0.5, and top-$k$\,=\,1024. We report run-to-run variation with the final measurements. Matched comparisons use identical timing boundaries and implementation settings. 

\section{Lossless Packing at \texorpdfstring{\(64\times\)}{64x} Spatial Compression}
\label{app:lossless_packing}

We evaluate whether our lossless packing strategy can increase the spatial
compression factor while preserving both tokenizer reconstruction and
text-to-image generation performance. Starting from
\(\mathrm{f32g8}\) (i.e., \(32\times\) spatial compression with \(G=8\);
other configurations follow the same notation), we pack each \(2\times2\)
neighborhood of spatial latent positions along the group dimension to obtain
\(\mathrm{f64g32}\). Specifically, four spatial positions with eight groups
each are rearranged into a single spatial position with 32 groups. This
operation is exactly invertible and neither discards nor re-quantizes any
latent codes. Consequently, applying the inverse packing operation before
decoding yields identical tokenizer reconstruction by construction, while
reducing the number of spatial latent positions by \(4\times\).

To assess the effect of packing on generation, we train the text-to-image
generator with the \(\mathrm{f64g32}\) representation using the same training
recipe as for the \(\mathrm{f32g8}\) baseline. The results are reported in
Table~\ref{tab:lossless_packing_generation}.

\begin{table}[t]
    \centering
    \caption{
        Text-to-image generation with lossless latent packing.
        The configuration with \(64\times\) spatial compression and \(G=32\)
        is obtained by losslessly packing the configuration with
        \(32\times\) spatial compression and \(G=8\), leaving tokenizer
        reconstruction unchanged. PRISM is the combined score on a
        0--10 scale.
    }
    \label{tab:lossless_packing_generation}
    \small
    \begin{tabular}{ccccc}
        \toprule
        Compression
        & \(G\)
        & GenEval \(\uparrow\)
        & DPG \(\uparrow\)
        & PRISM \(\uparrow\) \\
        \midrule
        \(32\times\) & 8  & 0.8060 & 82.49 & 5.30 \\
        \(64\times\) & 32 & 0.7885 & 80.44 & 5.13 \\
        \bottomrule
    \end{tabular}
\end{table}

As shown in Table~\ref{tab:lossless_packing_generation}, the packed
\(\mathrm{f64g32}\) representation achieves a GenEval score of 0.7885, a DPG
score of 80.44, and a PRISM score of 5.13. Compared with
\(\mathrm{f32g8}\), it retains strong generative performance despite using
a \(4\times\) shorter spatial latent sequence. These results demonstrate that
our lossless packing strategy enables effective text-to-image generation at
a \(64\times\) spatial compression factor without altering the reconstruction
capability of the tokenizer.

\clearpage
\section{Image Reconstruction Quality}
\label{app:reconstruction}
We also evaluate the reconstruction capability of our visual tokenizers on the ImageNet validation set at a fixed resolution of $512\times512$. We report peak signal-to-noise ratio (PSNR) and compare against five open-source visual tokenizers: VQGAN~\citep{esser2021taming}, IBQ~\citep{shi2025scalableimagetokenizationindex}, VAR~\citep{tian2024var}, Open-MAGVIT2~\citep{luo2025openmagvit2opensourceprojectdemocratizing}, and Emu3.5~\citep{cui2025emu35}.

As shown in Table~\ref{tab:reconstruction}, our \texttt{f32g16} tokenizer achieves the highest PSNR of 25.12\,dB at $32\times$ effective spatial compression, exceeding Open-MAGVIT2 by 1.14\,dB. At $64\times$ effective spatial compression, our \texttt{f64g32} variant achieves 23.90\,dB, outperforming VQGAN, IBQ, VAR, and Emu3.5 while remaining within 0.08\,dB of Open-MAGVIT2. These results show that our tokenizers retain competitive reconstruction fidelity at substantially higher effective spatial compression factors.

\begin{table}[htbp]
\centering
\small
\caption{Image reconstruction quality on the ImageNet validation set at $512\times512$ resolution. We report the effective per-axis spatial compression factor; for VAR it refers to the finest latent scale. Higher PSNR is better.}
\label{tab:reconstruction}
\begin{tabular}{lcc}
\toprule
Tokenizer & Spatial compression & PSNR (dB) $\uparrow$ \\
\midrule
VQGAN~\citep{esser2021taming} & $16\times$ & 22.24 \\
IBQ~\citep{shi2025scalableimagetokenizationindex} & $16\times$ & 23.39 \\
VAR~\citep{tian2024var} & $16\times$ & 23.43 \\
Open-MAGVIT2~\citep{luo2025openmagvit2opensourceprojectdemocratizing} & $16\times$ & 23.98 \\
Emu3.5~\citep{cui2025emu35} & $16\times$ & 23.82 \\
\midrule
GSTAR-Tokenizer (Ours) \texttt{f32g16} & $32\times$ & \textbf{25.12} \\
GSTAR-Tokenizer (Ours) \texttt{f64g32} & $64\times$ & 23.90 \\
\bottomrule
\end{tabular}
\end{table}

\begin{figure*}[htbp]
\centering
\includegraphics[width=\textwidth]{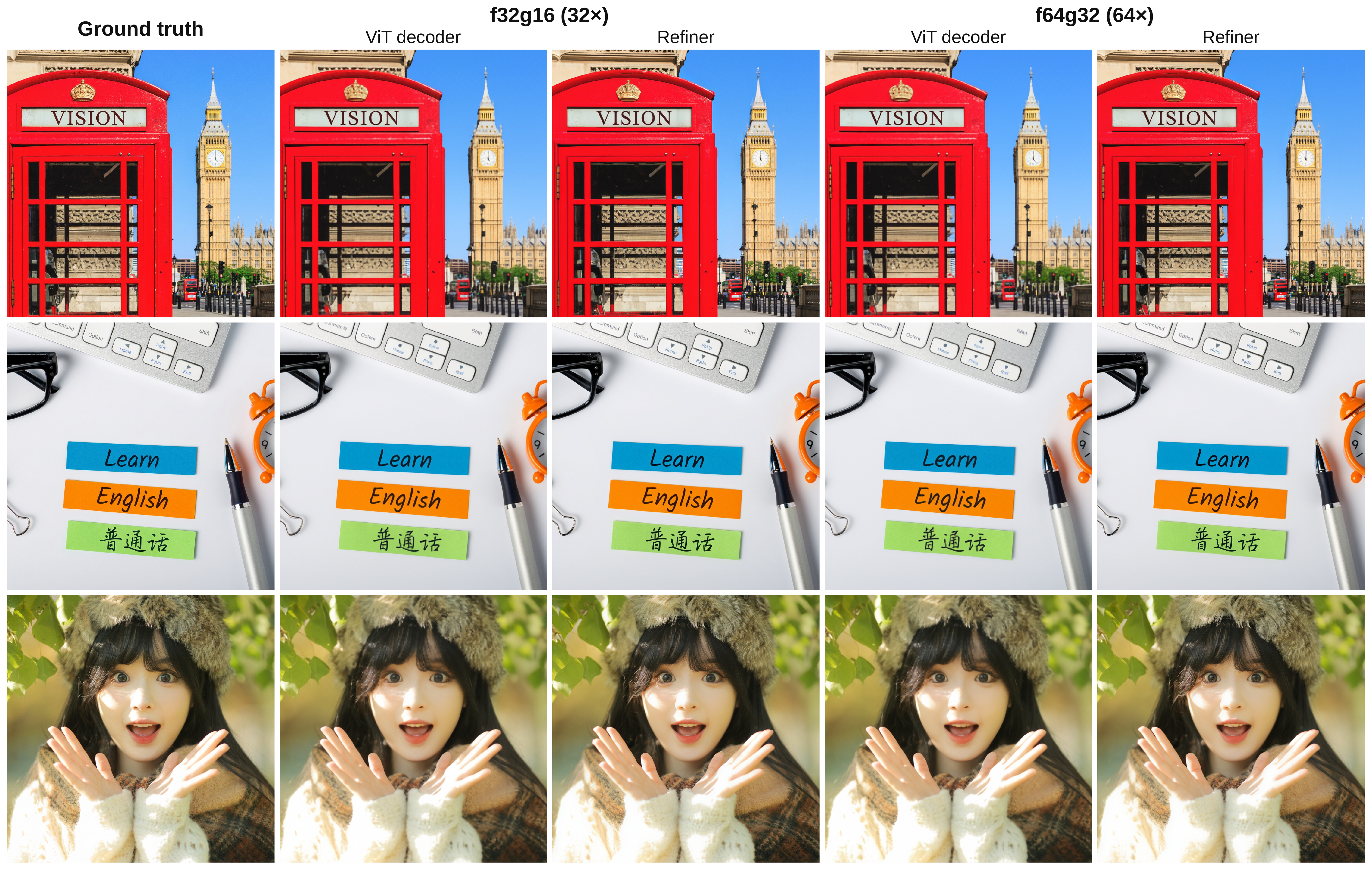}
\caption{Qualitative reconstruction examples. From left to right: ground truth, \texttt{f32g16} ViT decoder, \texttt{f32g16} refiner, \texttt{f64g32} ViT decoder, and \texttt{f64g32} refiner. Each row shows the same example across all methods. These visual examples are shown at $1024\times1024$ resolution, separately from the quantitative ImageNet evaluation at $512\times512$.}
\label{fig:refinercases}
\end{figure*}

We also present qualitative reconstruction examples in Figure~\ref{fig:refinercases}, comparing the ViT decoder and the refiner for both the \texttt{f32g16} and \texttt{f64g32} tokenizers. The examples illustrate how the refiner improves visual detail and sharpness while preserving the overall image content.

\end{document}